\documentclass{ecai}

\usepackage{latexsym}
\usepackage{amssymb}
\usepackage{amsmath}
\usepackage{amsthm}
\usepackage{booktabs}
\usepackage{enumitem}
\usepackage{graphicx}
\usepackage{color}
\usepackage{caption}

\newtheorem{theorem}{Theorem}

\newtheorem{proposition}{Proposition}

\newtheorem{definition}{Definition}

\usepackage{float}

\usepackage{algorithm}
\usepackage{algorithmic}
\usepackage[switch]{lineno}
 
\usepackage{color}
\usepackage[normalem]{ulem}
\usepackage{rotating}
\usepackage{tabularray}

\usepackage{graphicx}
\usepackage{subcaption}

\newlength\myindent
\newcommand{\BibTeX}{B\kern-.05em{\sc i\kern-.025em b}\kern-.08em\TeX}

\begin{document}


\begin{frontmatter}


\paperid{123} 


\title{Robust Time Series Forecasting with  Non-Heavy-Tailed \\Gaussian Loss-Weighted Sampler}


\author[A, B, C]{\fnms{Jiang}~\snm{YOU}
\thanks{Corresponding Author. Email: jiang.you@esiee.fr}}
\author[C]{\fnms{Arben}~\snm{CELA}}
\author[C]{\fnms{René}~\snm{NATOWICZ}}
\author[B]{\fnms{Jacob}~\snm{OUANOUNOU}}
\author[A]{\fnms{Patrick}~\snm{SIARRY}}

\address[A]{Université Paris-Est Créteil, Île-de-France, France}
\address[B]{HN-Services, Île-de-France, France}
\address[C]{ESIEE Paris, Île-de-France, France}


\begin{abstract}
Forecasting multivariate time series is a computationally intensive task challenged by extreme or redundant samples. Recent resampling methods aim to increase training efficiency by reweighting samples based on their running losses. However, these methods do not solve the problems caused by heavy-tailed distribution losses, such as overfitting to outliers. To tackle these issues, we introduce a novel approach: a Gaussian loss-weighted sampler that multiplies their running losses with a Gaussian distribution weight. It reduces the probability of selecting samples with very low or very high losses while favoring those close to average losses. As it creates a weighted loss distribution that is not heavy-tailed theoretically, there are several advantages to highlight compared to existing methods: 1) it relieves the inefficiency in learning redundant easy samples and overfitting to outliers, 2) It improves training efficiency by preferentially learning samples close to the average loss. Application on real-world time series forecasting datasets demonstrates improvements in prediction quality for $1\%$-$4\%$ using mean square error measurements in channel-independent settings. The code will be available online after the review. 
\end{abstract}

\end{frontmatter}


\section{Introduction}
Time series forecasting predicts future trends based on historical information. It has broad applications in forecasting traffic flow \citep{Lai_2018}, electricity transformer temperature \citep{zhou_fedformer_2022}, temperature and humidity in weather station \citep{liu2022pyraformer}. Recent advancements in this field have been driven by deep learning techniques, including transformer models \citep{liu2022pyraformer, Nie-2023-PatchTST} and convolutional neural networks \citep{wang2023timemixer, you_kun_2024}. 

Despite these advancements, challenges remain in training deep learning models on datasets with redundant samples and outliers. This issue, existing across the field, severely affects training efficiency and prediction accuracy. A notable factor contributing to this problem is that loss distributions often exhibit heavy tails \citep{heavy_2021}
 which degrades the performance of models. 
 
To address these challenges, researchers have developed various strategies, including reweighting the loss function with techniques like Focal Loss \citep{focal_loss_2020}, and Meta-Weight-Net 
\citep{han_2018_meta_weight_net}
, and data manipulation approaches such as over-sampling \citep{smote_2002}
, under-sampling \citep{undersampling_2020}, and data pruning \citep{qin2024infobatch}. These methods aim to mitigate issues stemming from the redundancy of simple samples and the presence of extreme outliers in datasets.

As a regression task, time series forecasting is particularly susceptible to the influence of outlier samples and the redundancy in learning from less informative majority samples. To the best of our knowledge, the application of these reweighting and resampling techniques specifically to time series datasets remains underexplored, except in Pyraformer\citep{liu2022pyraformer}, where the authors propose a static weight function that can increase the frequency of hard samples based on the amplitude of time series segment.

Moreover, existing solutions typically require class labels (Focal Loss \citep{focal_loss_2020}, SMOTE \citep{smote_2002}, unsuccessful to adequately address extreme values (Infobatch \citep{qin2024infobatch}), lack explicit weighting functions(Meta-Weight-Net \citep{han_2018_meta_weight_net}) or rely on algorithmic sample removal (under-sampling \citep{undersampling_2020}). These flaws limit their application in time series forecasting tasks.

\begin{figure*}[ht]
\centering
\includegraphics[width=\textwidth]{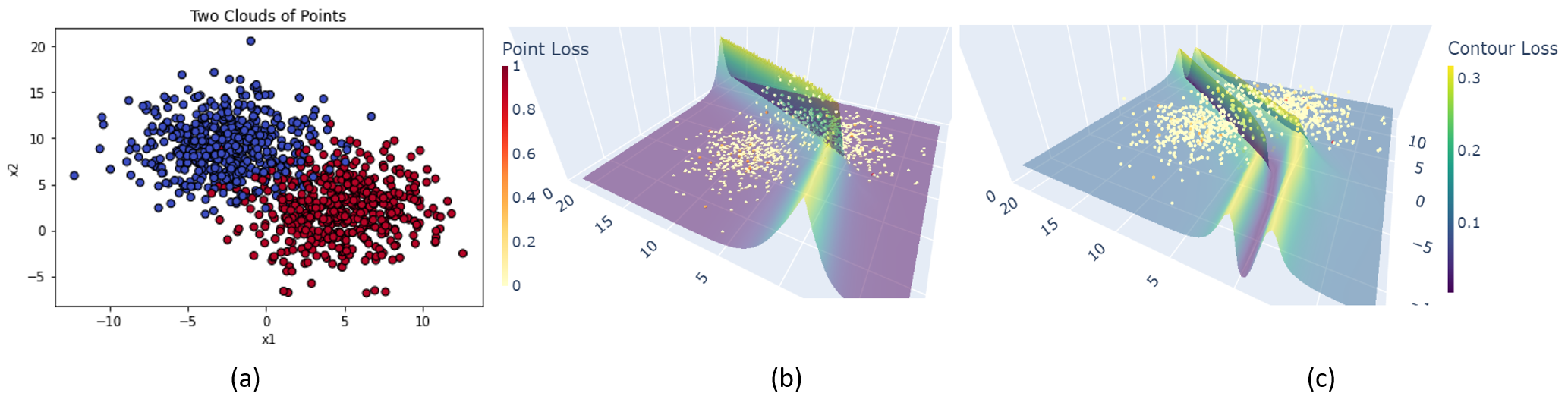}
\caption{(a) Synthetic dataset for the binary classification task. It contains 1000 samples of 2 classes following Gaussian distribution with standard deviation $\sigma=0.3$ and centered differently (b) The surface is the contour loss (The minimum distance between the predicted value and {0,1}) and the dot color represents point loss. (c) The surface is Gaussian weight($\mu$ = 0.0, $\sigma$ = 1.0) computed with the contour loss. It focuses on samples around the decision boundary but reduces the frequency of learning boundary (hard) samples or samples far away from the boundary (easy). Therefore, it is less affected by outliers and redundant samples. }
\label{fig:gaussian_loss_weighted_sampler}
\vspace{10pt}
\end{figure*}

To overcome these challenges, we propose a novel approach in this article: the Gaussian loss-weighted sampler. This method applies a Gaussian distribution as a weight to the running loss distribution of samples, reducing the selection probability of samples with excessively low or high losses and favoring those with losses close to the mean (Figure \ref{fig:gaussian_loss_weighted_sampler}). This intuitive approach aims to reduce training time and enhance the efficiency of learning from informative samples.

Moreover, this approach creates a reweighted loss distribution that is not heavy-tailed. Compared to existing methods, there are several advantages to highlight: 1) it has a clear theoretical limit when integrating its multiplication with the exponential term in a heavy-tailed distribution. 2), it relieves the inefficiency in learning redundant easy samples and overfitting to outliers, 3) It improves training efficiency by preferentially learning samples close to the average loss. 

Our experimental design involves testing the Gaussian loss-weighted resampling method on multiple time series forecasting datasets. These datasets are chosen to represent a range of complexities and real-world scenarios \citep{Lai_2018, zhou_fedformer_2022, liu2022pyraformer}. The setup is designed to compare the proposed method against contemporary baseline approaches, using mean square error metrics to evaluate performance and training epochs for efficiency. We choose methods such as InfoBatch for comparison. 

The Experimental result of Gaussian loss-weighted resampling to time series forecasting datasets yielded promising results, demonstrating $1\%-4\%$ improvements in accuracy and $10\%$ acceleration in training speed. These findings suggest that the method effectively addresses the challenges of heavy-tailed distributions, outliers, and sample redundancy, outperforming existing solutions in various cases. 

We summarize our contributions below.
\begin{itemize}
    \item We propose a theoretically non-heavy-tailed sampling method that addresses the issues of learning redundant easy samples and rare outliers in time series forecasting tasks.
    \item Our sampling algorithm features a simple mathematical formulation, simplifying the procedure compared to InfoBatch.
    \item It avoids learning outliers too often, thus improving prediction quality, while maintaining the same speedup for training efficiency as InfoBatch.
\end{itemize}

In conclusion, we proposed a Gaussian loss-weighted sampling method to tackle challenges such as sample redundancy and training efficiency. This is the first reweighting method that creates a non-heavy-tailed loss distribution theoretically. We observe the improvements in training efficiency and prediction quality. As this method is independent of the dataset and class labels, we are expecting another type of application in real-world datasets in the future.  

\section{Related Works}
Resampling methods have been employed in imbalanced learning for decades on tasks such as anomaly detection and outlier detection. As deep learning methods are sensible to the distribution of instances in the dataset, it raise difficulties in learning rare samples. Therefore, researchers in this field proposed methods such as oversampling and undersampling to improve the efficiency by manually modifying the sample distribution. More precisely, oversampling increases the population of minority instances, and downsampling removes random instances from the majority. Therefore, these methods could create a more balanced dataset and could increase training efficiency by relieving the bias naturally existing in the dataset.

\textbf{Oversampling (or Up-sampling)}: Oversampling enhances the diversity and number of instances in the minority class, thereby creating a more balanced dataset that can lead to improved model performance. Pioneering techniques such as Synthetic Minority Over-sampling Technique (SMOTE \citep{smote_2002}),  Adaptive Synthetic Sampling (ADASYN \citep{adasyn_2008}), LR-SMOTE \citep{liang_lr_smote_2020} and  Geometric SMOTE \citep{camacho_geometric_2022} are widely discussed in this domain. These methods generate new, synthetic data points by interpolating between existing minority class instances. These synthetic instances are not mere duplicates but are crafted to embody the subtle nuances of the minority class, enriching the dataset with valuable information for the learning algorithm to capture the complexity of the underrepresented class more effectively. 

\textbf{Undersampling (or Down-sampling)}: In contrast, undersampling addresses class imbalance by reducing the size of the majority class. Random undersampling(RUS),  is a straightforward approach that involves the removal of instances from the majority class, potentially leading to a more balanced dataset without the additional computational cost of generating new data. However, random removal can sometimes eliminate valuable information. To refine this process, algorithms such Tomek-Link (\citep{Tomek_1976,  class_using_tomek_link}), RUSBoost \citep{RUSBoost_2010}
, and ECUBoost \citep{Entropy_under_sampling_2020} have been developed. These algorithms aim to prune the dataset strategically by removing instances that are either redundant or that contribute to class overlap, thereby enhancing the efficiency and effectiveness of the training process. Through targeted removal, these undersampling strategies ensure that the resultant dataset maintains the integrity of the majority class while still addressing the imbalance.

While Oversampling and Undersampling require class labels, gradient-based resampling methods are independent of the class labels. In reinforcement learning, the Prioritized Experience Replay \citep{schaul2015prioritized}  method learns high-loss examples by sorting the whole dataset. Recent data pruning method InfoBatch demonstrates an approach to removing samples that have a smaller loss than the average. 

\textbf{Data pruning }
Data pruning techniques have evolved to streamline training datasets by eliminating less informative instances, thus accelerating training without sacrificing performance. The author in \citep{Raju2021AcceleratingDL} used the upper confidence bound to prune less valuable samples, focusing on those the model is uncertain about for maximum informational gain. 
\citep{Raju2021AcceleratingDL} 
The authors in \citep{He2023LargescaleDP} introduced a dynamic approach that adjusts to model needs by pruning based on sample uncertainty. Additionally, the InfoBatch \citep{qin2024infobatch} method speeds up training by randomly removing samples with losses smaller than the average, assuming they add little new knowledge. 

\textbf{Weighted loss}
Weighted loss functions address class distribution skew by emphasizing underrepresented classes, enhancing model focus on minority samples. Focal Loss\citep{focal_loss_2020} refines this by concentrating training on hard-to-classify examples, reducing majority class dominance in imbalanced datasets. Dual Focal Loss \citep{dual_focal_loss_2021} enhances this by considering both the ground truth and the next highest logit, improving confidence balance. Additionally, the Meta-Weight-Net \citep{han_2018_meta_weight_net} dynamically adjusts the weight function of running loss using an extra neural network.

In this article, we introduce a novel Gaussian loss-weighted sampler that utilizes a Gaussian distribution to weight sample losses, favoring those near the mean to enhance training efficiency and reduce selection of extreme loss samples. This approach not only speeds up training but also produces a non-heavy-tailed loss distribution, offering several advantages: it prevents learning inefficiencies from redundant samples and overfitting to outliers and maintains a theoretical limit that makes the reweighted loss not heavy-tailed. In the context of time series forecasting, our Gaussian loss-weighted sampling method sensitively chooses samples around average loss, outperforming recent pruning methods like InfoBatch and simplifying the algorithm with a bounded weight function.

\begin{figure}[tb]
\centering

\begin{subfigure}[t]{.475\columnwidth}
\includegraphics[width=\columnwidth]{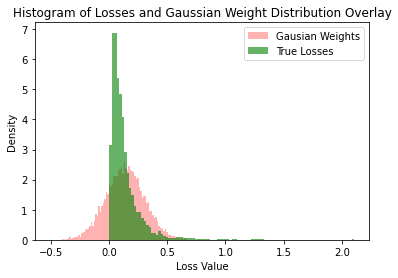} 
  \caption{True Loss and Gaussian weights.}
\end{subfigure}
 \hfill
\begin{subfigure}[t]{.475\columnwidth}
\includegraphics[width=\columnwidth]
{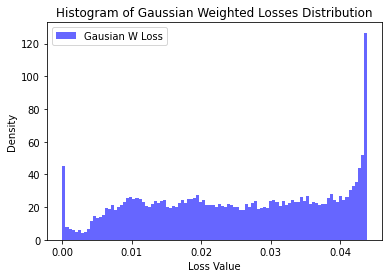}
  \caption{ Gaussian reweighted loss. ($\mu=-0.5$, $\sigma=1$)}
\end{subfigure}
\vspace{10pt}
  \caption{ Example of heavy-tailed true loss and reweighted loss from application of NLinear Method on ETTm2 dataset. The reweighted loss is more uniform.}
\label{fig:gaussian_loss_distribution}

\vspace{15pt}
 
\end{figure}

\section{Method}
In this section, we discuss Gaussian Loss-Weighted Sampling (GLWS), a method developed to improve the accuracy of time series forecasting models. First, we formulate the time series forecasting problem that makes predictions using historical information. Next, we explain the definition of weighted loss and resampling, which makes some data points more influential than others during model training. We then define the weight function by a Gaussian loss centered at the average value of running loss.  Lastly, we demonstrate that the gaussain weighed loss is not heavy-tailed which ensures its robustness against outliers.

In the following text, we note $z$ as the input data and $x$ as the loss obtained in training models. 

\subsection{Time series Forecasting}

We define the matrix $ z \in \mathbb{R}^{N \times M} $ as representing a multivariate time series dataset, where $ N $ is the dimension representing sampling time, and $ M $ represents the number of features. Let $ L $ denote the length of the memory or the look-back window. Then, the sequence $ (z_{t+1,1}, ..., z_{t+L,M}) $ (abbreviated as $ (z_{t+1}, ..., z_{t+L}) $) constitutes a segment of length $ L $ across all features, capturing historical data at time $ t $. The matrix $ Z_t \in \mathbb{R}^{L \times M} $ is termed the trajectory segment, representing this historical slice for each time $ t $ within the range $[0, N-L-1]$.

In time series forecasting, the dataset consists of a series of observed characteristics $ z $ and their corresponding future values $ \hat{z} $. Let $ z_t $ represent the feature at time step $ t $ and $ L $ be the length of the look-back window. The task involves using a historical series $ (z_{t+1}, ..., z_{t+L}) $ of length $ L $ to predict future values $ (\hat{z}_{t+L+1}, ..., \hat{z}_{t+L+T}) $ over the next $ T $ time steps. The fundamental problem in time series forecasting can thus be defined as:

$$
(\hat{z}_{t+L+1}, ..., \hat{z}_{t+L+T}) = F(z_{t+1}, ..., z_{t+L})
$$

Here, $ F $ is a predictive function that estimates the future values $ (\hat{z}_{t+L+1}, ..., \hat{z}_{t+L+T}) $ based on the input series $ (z_{t+1}, ..., z_{t+L}) $.

\subsection{Weighted loss and resampling}
 Let $X$ be a list of losses and $x\in X$, the expected weighted loss by resampling methods is:

\begin{eqnarray}\label{eq:def_resampling_discrete}
E(w, X) &=& \frac{1}{N}\sum_{x\in X}w(x)x \end{eqnarray}

where $N$ is the dataset size. The continuous version of this weighted loss is

\begin{eqnarray}\label{eq:def_resampling_continue}
E(w, X) & = & \int_{x\in X}w(x)f(x)x \end{eqnarray}

where the $w(x)$ is the weight function and $f(x)$ is the density function of loss of samples. 

Remark that statistically the expectation of the reweighted loss is the same as the expected loss obtained by resampling. In \cite{An2020WhyRO}, the author explains that resampling is more reliable than reweighting because reweighting amplifies the error gradient of minority outliers, making an algorithm harder to converge. Therefore, instead of reweighting the loss directly, we resample the data based on a custom distribution and then obtain the expected weighted loss.

In this work, we will discuss a Gaussian distribution centered on the average loss as the resampling weight function.

\subsection{Gaussian Loss Weighted Sampling}
This Gaussian Loss Weighted Sampler increases the number of learnable samples at each epoch of training iteration.  Therefore, it accelerates the training speed in time series forecasting. Gaussian Loss Weighted Sampler modifies the weight of each example based on their loss in real-time. This approach maintains the population of learnable samples at a larger size. As described in Algorithm \ref{alg:gaussian_loss_weighted_sampler} it dynamically computes and updates the weight assigned to each training sample, using the running loss as a basis. 

\begin{algorithm}[tb]
    
    \caption{Gaussian Loss Weighted Sampling}
    \label{alg:gaussian_loss_weighted_sampler}
    
    \textbf{Input}: dataset size $N$,  list of weights $w=[w_1,\ldots,w_n]$ and loss $x=[x_1,\ldots,x_n ]$,  user-defined bias $\mu$ and deviation $\sigma$.
    
    \begin{algorithmic} 
    \STATE \# Initialisation:
     \STATE Initiate $w$ as  a list of $1$ at the first run.
     \STATE \# Update weights of sampler
    \STATE Compute the mean $\mu_x$ and standard deviation $\sigma_x$   for $x$
    \STATE Normalization:  $y=\frac{x-\mu_x}{\sigma_x} $
     
    \STATE Compute weights:  $w=\frac{1}{\sigma\sqrt{2\pi}} e^{- \frac{1}{2\sigma^2 } (y-\mu)^2 }$
    \STATE update $w$ in sampler
    \end{algorithmic}
\end{algorithm}

Precisely, the algorithm recalibrates the weights based on the contribution of the sample to the overall loss. This ongoing adjustment ensures that samples causing higher losses or lower losses gain less attention in subsequent training batches, thereby saving more computing resources for samples around average losses, i.e., the samples not far from the decision boundary in the classification task or samples that are not too hard in regression tasks(Figure \ref{fig:gaussian_loss_distribution}). 

This computation is performed at the end of each epoch, allowing the weights to reflect recent performance metrics of the model. It leverages the inherent feedback loop within the training process, continuously refining the focus on problematic samples to optimize learning outcomes.

In our practical implementation of this sampling strategy, we utilize the \textit{WeightedRandomSampler} module from PyTorch. This module facilitates the resampling of data points for each training iteration based on custom-defined weights. By integrating this module, we can ensure that the training data fed into the model is not just randomly selected but is instead strategically chosen based on the evolving importance of each sample as determined by our algorithm. 

The random sampler is particularly useful in handling imbalanced datasets or in scenarios where certain classes or clusters of data are more challenging than others. In the context of forecasting , it allows more concentration on majority and informative samples in the dataset. The \textit{WeightedRandomSampler} thus plays a critical role in implementing our Gaussian Loss Weighted sampling approach effectively, enhancing the overall efficacy and efficiency of the model training process.

\subsection{Theory}
In this section, we demonstrate that the Gaussian loss-weighted loss forms a distribution that is not heavy-tailed. In the definition of heavy-tailed distribution, the integral of the product of an exponential term with such distribution towards infinity. Dramatically, by reweighting the loss with a Gaussian weight, this product becomes a biased Gaussian distribution with bias. therefore the integration is bounded by a constant. 
\\
\begin{definition}
Given a list of loss $x$, its distribution is heavy-tailed (right) if its density function$ f(x)$ verifies:

\begin{eqnarray}\label{eq:def_1}
\int_0^{+\infty}e^{\lambda x} f(x) &=&\infty\end{eqnarray}

\end{definition}

\begin{definition}
The Gaussian loss-weighted resampling loss is a multiplication of a Gaussian distribution and the loss,

\begin{eqnarray}\label{eq:def_2}
g(x)x &=&\frac{1}{\sigma\sqrt{2\pi}} e^{-  \frac{(y-\mu)^2}{2\sigma^2 }} x\end{eqnarray}

where $y=\frac{x-\mu_x}{\sigma_x}$  is normalized loss, $\mu_x$ and $\sigma_x$ are from distribution of $x$, $\mu$ and $\sigma$ are user-defined mean bias and deviation. 

\end{definition}

The bias $\mu$ and deviation $\sigma$ are user-defined hyper-parameters depending on their objectives and dataset. These parameters are borrowed from subjective probability and describe the preference of the user. Setting a negative $\mu$ means risk-averse, positive $\mu$ means risk-seeking and $\mu=0$ means risk neutral. Setting a $\sigma<1$ means selecting more samples around the mean loss and $\sigma>1$ means more uniform sampling (Figure \ref{fig:params_search}). By default, we use $\mu=0,\sigma=1$. 
\\

\begin{table*}[htb]
\caption{Multivariate time series forecasting results with Kernel U-Net and Gaussian Loss-Weighted Sampler. The prediction lengths $T$ is in \{96, 192, 336, 720\} and the lookback window $L$ is 720 for all datasets. We note the best results in \textbf{bold} and the second best results in \uline{underlined}.}
\label{tab:multivariate-table}

\vskip0.15in
\centering
\tiny
\begin{tblr}{
cell{1}{1}={c=2}{},
cell{1}{3}={c=2}{},
cell{1}{5}={c=2}{},
cell{1}{7}={c=2}{},
cell{1}{9}={c=2}{},
cell{1}{11}={c=2}{},
cell{1}{13}={c=2}{},
cell{1}{15}={c=2}{},
cell{1}{17}={c=2}{},
cell{2}{1}={c=2}{},
cell{3}{1}={r=4}{},
cell{7}{1}={r=4}{},
cell{11}{1}={r=4}{},
cell{15}{1}={r=4}{},
cell{19}{1}={r=4}{},
cell{23}{1}={r=4}{},
cell{27}{1}={r=4}{},
cell{38}{1}={c=2}{},
cell{38}{3}={c=2}{},
cell{38}{5}={c=2}{},
cell{38}{7}={c=2}{},
cell{38}{9}={c=2}{},
cell{38}{11}={c=2}{},
cell{38}{13}={c=2}{},
cell{38}{15}={c=2}{},
cell{39}{1}={c=2}{},
cell{41}{1}={r=4}{},
cell{45}{1}={r=4}{},
cell{49}{1}={r=4}{},
cell{53}{1}={r=4}{},
vline{2,3,4,5,6,7,8,9,10,12,14,11,13,15,17}={1-31}{},
hline{1-3,7,11,15,19,23,27,31}={-}{},
rowsep=0.35ex,colsep=3.5ex
}
Methods&&K-U-Net(+Gaussian)&&K-U-Net&&PatchTST&&Nlinear&&Dlinear&&FEDformer&&Autoformer&&Informer&\\
Metric&&MSE&MAE&MSE&MAE&MSE&MAE&MSE&MAE&MSE&MAE&MSE&MAE&MSE&MAE&MSE&MAE\\
\begin{sideways}ETTh1\end{sideways}&96&\textbf{0.366}&\textbf{0.392}&\uline{0.370}&\uline{0.393}&0.37&0.4&0.374&0.394&0.375&0.399&0.376&0.419&0.449&0.459&0.865&0.713\\
&192&\textbf{0.397}&\uline{0.415}&\uline{0.404}&\textbf{0.414}&0.413&0.429&0.408&0.415&0.405&0.416&0.42&0.448&0.5&0.482&1.008&0.792\\
&336&\textbf{0.417}&\uline{0.433}&\uline{0.420}&\textbf{0.430}&0.422&0.44&0.429&0.427&0.439&0.443&0.459&0.465&0.521&0.496&1.107&0.809\\
&720&\textbf{0.433}&\textbf{0.451}&\uline{0.438}&\uline{0.454}&0.447&0.468&0.44&0.453&0.472&0.49&0.506&0.507&0.514&0.512&1.181&0.865\\
\begin{sideways}ETTh2\end{sideways}&96&\textbf{0.269}&\textbf{0.332}&\uline{0.271}&\uline{0.335}&0.274&0.337&0.277&0.338&0.289&0.353&0.346&0.388&0.358&0.397&3.755&1.525\\
&192&\textbf{0.330}&\textbf{0.375}&\uline{0.332}&\uline{0.377}&0.339&0.379&0.344&0.381&0.383&0.418&0.429&0.439&0.456&0.452&5.602&1.931\\
&336&\uline{0.356}&\uline{0.395}&0.357&0.4&\textbf{0.329}&\textbf{0.384}&0.357&0.4&0.448&0.465&0.496&0.487&0.482&0.486&4.721&1.835\\
&720&\uline{0.382}&\uline{0.443}&0.39&0.438&\textbf{0.379}&\textbf{0.422}&0.394&0.436&0.605&0.551&0.463&0.474&0.515&0.511&3.647&1.625\\
\begin{sideways}ETTm1\end{sideways}&96&\textbf{0.281}&\textbf{0.340}&\uline{0.286}&\uline{0.342}&0.29&0.342&0.306&0.348&0.299&0.343&0.379&0.419&0.505&0.475&0.672&0.571\\
&192&\textbf{0.323}&\textbf{0.367}&\uline{0.330}&\uline{0.363}&0.332&0.369&0.349&0.375&0.335&0.365&0.426&0.441&0.553&0.496&0.795&0.669\\
&336&\textbf{0.352}&\textbf{0.384}&\uline{0.360}&\uline{0.384}&0.366&0.392&0.375&0.388&0.369&0.386&0.445&0.459&0.621&0.537&1.212&0.871\\
&720&\textbf{0.398}&\textbf{0.412}&\uline{0.405}&\uline{0.412}&0.416&0.42&0.433&0.422&0.425&0.421&0.543&0.49&0.671&0.561&1.166&0.823\\
\begin{sideways}ETTm2\end{sideways}&96&\textbf{0.158}&\textbf{0.248}&\uline{0.16}&\uline{0.245}&0.165&0.255&0.167&0.255&0.167&0.26&0.203&0.287&0.255&0.339&0.365&0.453\\
&192&\textbf{0.213}&\textbf{0.287}&\uline{0.215}&\uline{0.215}&0.22&0.292&0.221&0.293&0.224&0.303&0.269&0.328&0.281&0.34&0.533&0.563\\
&336&\textbf{0.265}&\textbf{0.323}&\uline{0.268}&\uline{0.326}&0.274&0.329&0.274&0.327&0.281&0.342&0.325&0.366&0.339&0.372&1.363&0.887\\
&720&\textbf{0.339}&\textbf{0.376}&\uline{0.343}&\uline{0.379}&0.362&0.385&0.368&0.384&0.397&0.421&0.421&0.415&0.433&0.432&3.379&1.338\\
\begin{sideways}Electricity\end{sideways}&96&\textbf{0.127}&\uline{0.226}&\uline{0.129}&0.226&0.129&\textbf{0.222}&0.141&0.237&0.14&0.237&0.193&0.308&0.201&0.317&0.274&0.368\\
&192&\textbf{0.145}&\uline{0.244}&\uline{0.147}&0.244&0.147&\textbf{0.24}&0.154&0.248&0.153&0.249&0.201&0.315&0.222&0.334&0.296&0.386\\
&336&\textbf{0.161}&\uline{0.261}&\uline{0.163}&0.261&0.163&\textbf{0.259}&0.171&0.265&0.169&0.267&0.214&0.329&0.231&0.338&0.3&0.394\\
&720&\textbf{0.195}&\uline{0.292}&\uline{0.197}&0.292&0.197&\textbf{0.29}&0.21&0.297&0.203&0.301&0.246&0.355&0.254&0.361&0.373&0.439\\
\begin{sideways}Traffic\end{sideways}&96&\textbf{0.356}&0.253&\uline{0.358}&\uline{0.253}&0.36&\textbf{0.249}&0.41&0.279&0.41&0.282&0.587&0.366&0.613&0.388&0.719&0.391\\
&192&\textbf{0.370}&0.262&\uline{0.373}&\uline{0.262}&0.379&\textbf{0.25}&0.423&0.284&0.423&0.287&0.604&0.373&0.616&0.382&0.696&0.379\\
&336&\textbf{0.390}&0.271&\textbf{0.390}&\uline{0.271}&0.392&\textbf{0.264}&0.435&0.29&0.436&0.296&0.621&0.383&0.622&0.337&0.777&0.42\\
&720&\textbf{0.430}&0.292&\textbf{0.430}&\uline{0.292}&0.432&\textbf{0.286}&0.464&0.307&0.466&0.315&0.626&0.382&0.66&0.408&0.864&0.472\\
\begin{sideways}Weather\end{sideways}&96&\textbf{0.140}&\uline{0.195}&\uline{0.142}&0.195&0.149&0.198&0.182&0.232&0.176&0.237&0.217&0.296&0.266&0.336&0.3&0.384\\
&192&\textbf{0.185}&\uline{0.244}&\uline{0.188}&0.244&0.194&0.241&0.225&0.269&0.22&0.282&0.276&0.336&0.307&0.367&0.598&0.544\\
&336&\textbf{0.239}&\uline{0.284}&\uline{0.241}&0.284&0.245&0.282&0.271&0.301&0.265&0.319&0.339&0.38&0.359&0.395&0.578&0.523\\
&720&\textbf{0.309}&\uline{0.333}&\uline{0.310}&0.333&0.314&0.334&0.338&0.348&0.323&0.362&0.403&0.428&0.419&0.428&1.059&0.741
\end{tblr}

\end{table*}

\begin{proposition}
Given a loss $x$ whose density function $f(x)$ is lipschitzien continue, there is a constant $C$ such that Gaussian loss-weighted resampling loss forms a distribution whose density function is $g(x)f(x)$ and 

\begin{eqnarray}\label{eq:propo_1}
C \int_{-\infty}^{+\infty}g(x)f(x) & = & 1\end{eqnarray}
\end{proposition}

\textbf{Proof}

1) The multiplication $g(x)f(x)$ is  positive because by definition $g(x)$ and $f(x)$ are positive:
$$g(x)f(x) > 0$$
Consequently, 

\begin{eqnarray}
\label{eq:proof_1} 
\int_{-\infty}^{+\infty}g(x)f(x) > 0\end{eqnarray}
2), There is a constant $C$ that the integral of new weigted loss can be scaled to $1$. The integral of multiplication of Gaussian weights and density function of loss is bounded by $1$:
$$\int_{-\infty}^{+\infty}g(x)f(x) \leq \int_{-\infty}^{+\infty}g(x)\leq 1$$

As $f(x)$ is a density function then, 
$$\exists t>0, 1 > f(t) > 0$$
Suppose the density function $f(x)$ is lipchizien continue then :
$\exists \Delta t > 0$ and $M > 0$, such that  
 $$\int_{t - \Delta t}^{t + \Delta t}g(x)f(x) \geq M$$

Therefore, 
$$ \exists  1>M>0,\int_{-\infty}^{+\infty}g(x)f(x) \geq M$$

Therefore there is a constant C that can scale the new weigted loss to be scaled to 1:
\begin{eqnarray}
\label{eq:proof_2} 
\exists  1\leq C\leq \frac{1}{M},  C\int_{-\infty}^{+\infty}g(x)f(x) =1
\end{eqnarray}
In this case, $Cg(x)f(x)$ forms a density function of a distribution.

3), This integral is $\sigma$-additive because the product $g(x)f(x)$ is positive and its integral is monotonically non-decreasing when expanding a segment to its neighbors.

\begin{eqnarray}
\label{eq:proof_3} 
\forall t\in R,\int_{t - \Delta t}^{t + \Delta t}g(x)f(x) \leq \int_{t - 2\Delta t}^{t + 2\Delta t}g(x)f(x) \end{eqnarray}

With this short proof in $1),2),3),$ we have a constant $C \in[1, \frac{1}{M}]$ that $Cg(x)f(x)$ is a density function of a distribution.

\begin{theorem}
The scaled weighted loss distribution which has a density function  $C g(x) f(x)$ is not heavy-tailed:

\begin{eqnarray}\label{eq:def_3}
\exists  C' , \int_0^{+\infty}e^{\lambda x } Cg(x)f(x) & \leq & C'\end{eqnarray}

where $C'$ is the upper bound of this integral.

\end{theorem}
\vspace{10pt}
\textbf{Proof:}
Suppose f a density function of loss that verified the definition of heavy-tailed(right) distribution: 

$$\int_0^{+\infty}e^{\lambda x }f(x) =\infty$$

As the density function is bounded by 1, we have
$$f(x)<1\leq   e^{\lambda x}, \forall x>0.$$
 This condition is verified because the loss $x$ is in general positive.

We multiply the Gaussian weights with running loss and we have :

$$\int_0^{+\infty}e^{\lambda x }C g(x)f(x) $$
$$=C\int_0^{+\infty}\frac{1}{\sigma\sqrt{2\pi}} e^{\lambda x }e^{-  \frac{(y-\mu)^2}{2\sigma^2 }} f(x)$$

$$<C\int_0^{+\infty}\frac{1}{\sigma\sqrt{2\pi}} e^{\lambda x }e^{-   \frac{(y-\mu)^2}{2\sigma^2 }} e^{\lambda x }$$

$$=C\int_0^{+\infty}\frac{1}{\sigma\sqrt{2\pi}} e^{-   \frac{(y-\mu)^2}{2\sigma^2}  +2\lambda x} $$

$$=C\int_0^{+\infty}\frac{1}{\sigma\sqrt{2\pi}} e^{-  \frac{(x-\mu_x-\mu\sigma_x )^2}{2\sigma_x^2 \sigma^2 }+2\lambda x} $$

$$=C'\int_0^{+\infty} \frac{1}{B\sqrt{2\pi}} e^{- \frac{(x-A)^2}{2B^2}} $$

$$\leq C'$$

Where$ A, B, C, C' $are constants computed from $(\mu,\sigma ,\mu_x ,\sigma_x,\lambda) .$ Please check Appendix for detailed computation. 

Remark that the term in the last step is a Gaussian distribution so it is bounded by 1. Therefore the entire integral is bounded by $C'$

The Gaussian weighted loss is a new distribution that makes the integral of its multiplication with $e^{\lambda x }$bounded, thus making it no longer a heavy-tailed distribution.

\section{Experiments and Result}

In this section, we conduct experiments to demonstrate the efficacy of Gaussian sampling techniques on recent time series forecasting models, specifically focusing on improvements in accuracy and training efficiency.

Furthermore, we compare the performance of the Gaussian loss-weighted sampler with two prominent methods: InfoBatch and Meta-Weight Net. These comparisons are carried out using the K-U-Net architecture on the ETT dataset. The Gaussian loss-weighted sampler enhances learning efficiency by focusing on informative samples.

Additionally, we extend our evaluation to include the application of the Gaussian loss-weighted sampler on PatchTST\citep{Nie-2023-PatchTST} and CARD\citep{xue2024card} models to enrich the results and provide a comprehensive view of how Gaussian sampling influences different types of neural network architectures. These comparisons aim to highlight the generality and potential benefits of incorporating Gaussian sampling into various deep-learning frameworks.

\begin{table*}[hbt]
\caption{ Application of Gaussian Loss-Weighted Resampling, Infobatch on Kernel U-Net. We note the best results in \textbf{bold} and the second best results in \uline{underlined}}
\label{tab:multivariate-comparison-infobatch}

\vspace{5pt}
\tiny
\centering

\begin{tblr}{
cell{1}{1}={c=2}{},
cell{1}{3}={c=2}{},
cell{1}{5}={c=2}{},
cell{1}{7}={c=2}{},
cell{1}{9}={c=2}{},
cell{1}{11}={c=2}{},
cell{1}{13}={c=2}{},
cell{1}{15}={c=2}{},
cell{1}{17}={c=2}{},
cell{2}{1}={c=2}{},
cell{3}{1}={r=4}{},
cell{7}{1}={r=4}{},
cell{11}{1}={r=4}{},
cell{15}{1}={r=4}{},
vline{2,3,4,5,6,7,8,9,10,12,14,11,13,15,17}={1-31}{},
hline{1-3}={-}{},
hline{7,11,15,19}={1-18}{},
rowsep=0.4ex,colsep=2.5ex
}
Method&&{K-U-Net\\Linear\_1111}&&{K-U-Net\\Linear\_1111\\(+Gaussian)}&&{K-U-Net\\Linear\_1111\\(+IB(s=0.3))}&&{K-U-Net\\Linear\_1111\\(+IB(s=0.5))}&&{K-U-Net\\Linear\_0000}&&{K-U-Net\\Linear\_0000\\(+Gaussian)}&&{K-U-Net\\Linear\_0000\\(+IB(s=0.3))}&&{K-U-Net\\Linear\_0000\\(+IB(s=0.5))}&\\
Metric&&MSE&{Top5-\\MMSE}&MSE&{Top5-\\MMSE}&MSE&{Top5-\\MMSE}&MSE&{Top5-\\MMSE}&MSE&{Top5-\\MMSE}&MSE&{Top5-\\MMSE}&MSE&{Top5-\\MMSE}&MSE&{Top5-\\MMSE}\\
\begin{sideways}ETTh1\end{sideways}&96&\uline{0.370}&\uline{0.369}&\textbf{0.365}&\textbf{0.365}&0.377&0.375&0.385&0.380&\textbf{0.372}&\textbf{0.371}&\uline{0.372}&\uline{0.372}&0.372&0.372&0.374&0.373\\
&192&\uline{0.410}&\uline{0.404}&\textbf{0.403}&\textbf{0.401}&0.418&0.409&0.420&0.416&\uline{0.409}&\uline{0.411}&\textbf{0.408}&\textbf{0.411}&0.411&0.413&0.414&0.414\\
&336&0.426&\textbf{0.421}&\textbf{0.424}&0.422&0.445&0.426&0.446&0.430&\uline{0.441}&\uline{0.442}&\textbf{0.440}&\textbf{0.441}&0.442&0.442&0.444&0.442\\
&720&\uline{0.472}&\uline{0.465}&\textbf{0.452}&\textbf{0.448}&0.504&0.474&0.497&0.490&\uline{0.456}&\uline{0.454}&\textbf{0.442}&\textbf{0.441}&0.462&0.457&0.465&0.457\\
\begin{sideways}ETTh2\end{sideways}&96&0.311&0.291&\textbf{0.288}&\textbf{0.282}&\uline{0.321}&\uline{0.294}&0.326&0.297&\textbf{0.271}&\textbf{0.271}&0.274&0.272&\uline{0.272}&\uline{0.272}&0.273&0.273\\
&192&0.396&0.380&\textbf{0.360}&\textbf{0.370}&0.400&0.376&\uline{0.392}&\uline{0.377}&\textbf{0.334}&\textbf{0.352}&0.345&0.359&\uline{0.335}&\uline{0.352}&0.335&0.352\\
&336&0.421&0.388&\textbf{0.381}&\textbf{0.384}&\uline{0.407}&\uline{0.386}&0.416&0.388&\textbf{0.357}&\textbf{0.360}&0.376&0.375&\uline{0.358}&\uline{0.359}&0.359&0.357\\
&720&0.472&0.429&\textbf{0.444}&\textbf{0.421}&\uline{0.458}&\uline{0.433}&0.476&0.431&\textbf{0.386}&\textbf{0.386}&0.391&0.391&\uline{0.387}&\uline{0.386}&0.387&0.386\\
\begin{sideways}ETTm1\end{sideways}&96&\uline{0.290}&\uline{0.287}&\textbf{0.288}&\textbf{0.285}&0.298&0.288&0.293&0.290&\uline{0.308}&\uline{0.304}&\textbf{0.306}&\textbf{0.303}&0.309&0.304&0.309&0.305\\
&192&\uline{0.330}&\uline{0.329}&\textbf{0.327}&\textbf{0.327}&0.333&0.331&0.334&0.333&\uline{0.337}&\uline{0.336}&\textbf{0.335}&\textbf{0.334}&0.337&0.336&0.337&0.336\\
&336&\uline{0.359}&\uline{0.359}&\textbf{0.355}&\textbf{0.356}&0.363&0.362&0.365&0.365&\uline{0.366}&\uline{0.365}&\textbf{0.364}&\textbf{0.364}&0.368&0.366&0.366&0.366\\
&720&\uline{0.411}&\uline{0.407}&\textbf{0.403}&\textbf{0.402}&0.411&0.409&0.417&0.411&\uline{0.416}&\uline{0.415}&\textbf{0.415}&\textbf{0.415}&0.416&0.416&0.418&0.416\\
\begin{sideways}ETTm2\end{sideways}&96&0.166&0.163&\textbf{0.159}&\textbf{0.159}&0.168&0.164&0.172&0.167&\uline{0.162}&\uline{0.161}&\textbf{0.158}&\textbf{0.158}&0.162&0.162&0.162&0.162\\
&192&0.233&\uline{0.222}&\textbf{0.222}&\textbf{0.216}&0.230&0.223&\uline{0.228}&0.225&\uline{0.216}&\uline{0.215}&\textbf{0.214}&\textbf{0.213}&0.217&0.216&0.217&0.217\\
&336&0.282&0.273&\textbf{0.270}&\textbf{0.265}&0.276&\uline{0.273}&\uline{0.275}&0.276&\uline{0.268}&\uline{0.267}&\textbf{0.266}&\textbf{0.265}&0.269&0.267&0.269&0.268\\
&720&0.389&0.353&\textbf{0.346}&\textbf{0.348}&0.359&\uline{0.356}&\uline{0.350}&0.357&\uline{0.350}&\uline{0.352}&\textbf{0.349}&\textbf{0.351}&\textbf{0.349}&0.353&0.350&0.353
\end{tblr}

\end{table*}

\subsection{Datasets}. 

We conducted experiments with recent Kernel U-Net on 7 public datasets, including 4 ETT (Electricity Transformer Temperature) datasets\footnote{https://github.com/zhouhaoyi/ETDataset} (ETTh1, ETTh2, ETTm1, ETTm2), Weather\footnote{https://www.bgc-jena.mpg.de/wetter/}, Traffic\footnote{http://pems.dot.ca.gov} and Electricity\footnote{https://archive.ics.uci.edu/ml/datasets/ElectricityLoadDiagrams \\
20112014}.

The datasets have been benchmarked and publicly available on \citep{wu_autoformer_2021}. We refer the description of dataset from \citep{Zeng_AreTE_2022}.
\begin{itemize}
    \item Electricity Transformer Temperature (ETT) dataset comprises 2 years of 2 electricity transformers collected from 2 stations, including the oil temperature and 6 power load features. Observations every hour (i.e., ETTh1) and every 15 minutes (i.e., ETTm1) are provided. 
    \item Weather includes 21 indicators of weather, such as air temperature, and humidity. Its data is recorded every 10 min for 2020 in Germany.
    \item Traffic describes the road occupancy rates. It contains the hourly data recorded by the sensors of San Francisco freeways from 2015 to 2016.
    \item Electricity collects the hourly electricity consumption of 321 clients from 2012 to 2014.
\end{itemize}

Here, we followed the experiment setting in \citep{Zeng_AreTE_2022} and partitioned the data into 12, 4, and 4 months for training, validation, and testing respectively for the ETT dataset. The data is split into $[0.7,0.1,0.2]$ for training, validation, and testing for the dataset Weather, Traffic, and Electricity.

In the time series forecasting task, we follow the experiment setting in NLinear \citep{Zeng_AreTE_2022}, which takes $L={720}$ step historical data as input and forecast $T\in{96,192,336,720}$ step value in the future. We use mean value normalization and instance normalization for the ETT dataset.

We include five recent SOTA methods: PatchTST \citep{Nie-2023-PatchTST}, Kernel-U-Net\citep{you_kun_2024}, NLinear and DLinear\citep{Zeng_AreTE_2022}, FEDFormer \citep{zhou_fedformer_2022}, AutoFormer\cite{wu_autoformer_2021}, Informer\citep{haoyietal-informer-2021}. We merge the results reported in \citep{Nie-2023-PatchTST, Zeng_AreTE_2022, you_kun_2024} to take their best report.

Following previous works \citep{wu_autoformer_2021}, we use Mean Squared Error (MSE) and Mean Absolute Error (MAE) as the core metrics to compare performance for Forecasting problems. 

We set lookback window $L=720$ in experiments. The list of multiples for kernel-u-net is respectively [5,6,6] and the segment-unit input length is 4. The hidden dimensions are 128 for multivariate tasks. The hidden dimension for CARD is 16 and 128 for PatchTST. We take the default setting in their original paper. The input dimension is 1 as we follow the channel-independent setting in \citep{Zeng_AreTE_2022}  and \citep{Nie-2023-PatchTST}. We use the Random Erase \citep{zhong2020random} from image processing for data augmentation. The learning rate is selected in [0.00005, 0.0001] depending on the dataset. The training epoch is 100 and the patience is 20 in general. 

We apply weighted early stopping where the triggering MSE (TMSE) \citep{you_kun_2024} is computed with 
$TMSE_{e+1} = \tau * TMSE_{e} + (1-\tau) * VMSE_e$
, where $\tau$ is a hyper-parameter to be set, $e$ is the index of the episode, and VMSE is the running MSE on a validation set. The parameter $\tau$ sets the sensitivity of early stopping and is empirically selected. In general, it is set to be 0.9 or 0.5 in the fine-tuning stage. 

The MSE loss function is defined with $: \mathcal{L}=\mathbb{E}_{\boldsymbol{x}} \frac{1}{B*M} \sum_{i=1}^{B*M}\left\|\hat{\boldsymbol{x}}_{L+1: L+T}^{(i)}-\boldsymbol{x}_{L+1: L+T}^{(i)}\right\|_2^2$ , where $B$ is the batch size and $M$ is the multiple of feature size unit of time series.

we also report results with the Top5 running Minimum MSE (Top5-MMSE) in the ablation study.

\subsection{Result}
\textbf{Main result.}

The Gaussian Loss-Weighted sampler has improvements in mean squared error (MSE) scores with K-U-Net. The improvement of MSE is in average $1.4\%$ (Table \ref{tab:multivariate-table}).  We show the fintune result with the selected variant of K-U-Net.

\begin{figure}[hbt]
\centering

\begin{subfigure}[b]{.475\columnwidth}
\includegraphics[width=\columnwidth]{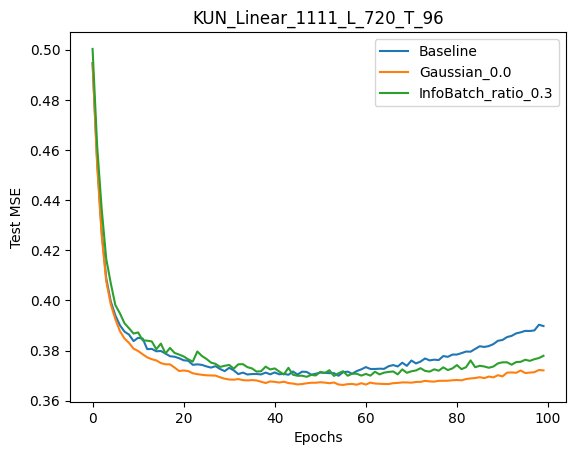} 
  \caption{ETTh1, L=720, T=96}
\end{subfigure}
 \hfill
\begin{subfigure}[b]{.475\columnwidth}
\includegraphics[width=\columnwidth]
{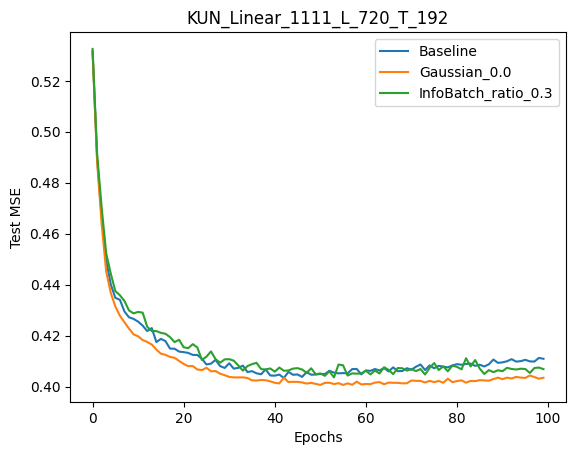}
  \caption{ETTh1, L=720, T=192}
\end{subfigure}

\vspace{10pt}

\begin{subfigure}[b]{.475\columnwidth}
\includegraphics[width=\columnwidth]
{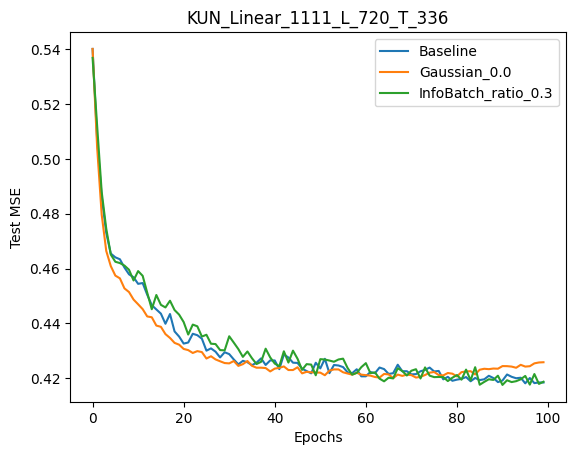}
  \caption{ETTh1, L=720, T=336}
\end{subfigure}
 \hfill
\begin{subfigure}[b]{.475\columnwidth}
\includegraphics[width=\columnwidth]
{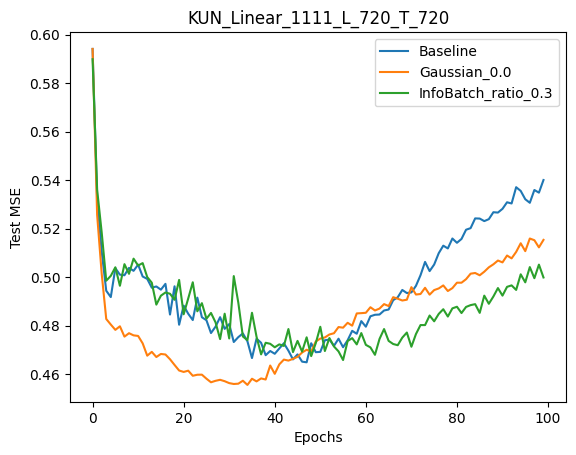}
  \caption{ETTh1, L=720, T=720}
\end{subfigure}

\vspace{10pt}
\caption{Running test MSE of Gaussian Method and Infobatch with pruning ratio $\sigma $=$ 0.3$ and baseline without resampling.}
\label{fig:acceleration}
\vspace{15pt}
\end{figure}

\textbf{Acceleration}

We measure the number of epochs to reach the best score of the baseline method. in general, the Gaussian Loss weighted sampler can reduce around $10\%$ training epochs to reach the optimum of the baseline method (Figure \ref{fig:acceleration}).

\begin{table}[htb]
\caption{ Application of Gaussian Loss Weighted Sampler on CARD and PatchTST.}
\label{tab:multivariate-abalation_card_patchtst}

\vspace {10pt}
\tiny
\centering
\begin{tblr}{
cell{1}{1}={c=2}{},
cell{1}{3}={c=2}{},
cell{1}{5}={c=2}{},
cell{1}{7}={c=2}{},
cell{1}{9}={c=2}{},
cell{2}{1}={c=2}{},
cell{3}{1}={r=4}{},
cell{7}{1}={r=4}{},
cell{11}{1}={r=4}{},
cell{15}{1}={r=4}{},
vline{3,5,7,9}={1-2}{},
vline{2,3,5,7,9}={3-18}{},
hline{1-3,7,11,15,19}={1-10}{},
rowsep=0.15ex,colsep=2.5ex
}
Method&&CARD&&{CARD\\(+Gaussian)}&&PatchTST&&{PatchTST\\(+Gaussian)}&\\
Metric&&MSE&{Top5-\\MMSE}&MSE&{Top5-\\MMSE}&MSE&{Top5-\\MMSE}&MSE&{Top5-\\MMSE}\\
\begin{sideways}ETTh1\end{sideways}&96&0.378&0.377&\textbf{0.367}&\textbf{0.366}&0.387&0.380&\textbf{0.378}&\textbf{0.377}\\
&192&0.408&0.407&\textbf{0.399}&\textbf{0.398}&\textbf{0.433}&\textbf{0.419}&\textbf{0.433}&0.422\\
&336&0.425&0.419&\textbf{0.424}&\textbf{0.417}&0.461&\textbf{0.444}&\textbf{0.450}&\textbf{0.444}\\
&720&0.429&\textbf{0.419}&\textbf{0.421}&0.421&\textbf{0.465}&0.465&\textbf{0.465}&\textbf{0.458}\\
\begin{sideways}ETTh2\end{sideways}&96&\textbf{0.273}&\textbf{0.271}&0.274&0.274&0.300&0.284&\textbf{0.272}&\textbf{0.270}\\
&192&\textbf{0.336}&\textbf{0.344}&\textbf{0.336}&0.351&0.354&0.367&\textbf{0.348}&\textbf{0.363}\\
&336&0.370&0.366&\textbf{0.347}&\textbf{0.349}&\textbf{0.372}&0.372&\textbf{0.372}&\textbf{0.369}\\
&720&0.394&0.395&\textbf{0.386}&\textbf{0.384}&0.395&\textbf{0.401}&\textbf{0.393}&\textbf{0.401}\\
\begin{sideways}ETTm1\end{sideways}&96&0.291&0.289&\textbf{0.285}&\textbf{0.285}&0.295&0.291&\textbf{0.290}&\textbf{0.288}\\
&192&0.334&0.333&\textbf{0.327}&\textbf{0.327}&0.334&0.335&\textbf{0.326}&\textbf{0.322}\\
&336&0.369&0.368&\textbf{0.356}&\textbf{0.356}&0.377&0.376&\textbf{0.357}&\textbf{0.355}\\
&720&0.452&0.416&\textbf{0.410}&\textbf{0.409}&0.430&0.431&\textbf{0.418}&\textbf{0.418}\\
\begin{sideways}ETTm2\end{sideways}&96&0.169&0.166&\textbf{0.160}&\textbf{0.159}&0.175&0.166&\textbf{0.163}&\textbf{0.160}\\
&192&0.228&0.224&\textbf{0.214}&\textbf{0.211}&0.224&0.224&\textbf{0.220}&\textbf{0.220}\\
&336&0.277&0.298&\textbf{0.271}&\textbf{0.271}&0.268&0.277&\textbf{0.268}&\textbf{0.277}\\
&720&\textbf{0.366}&0.380&\textbf{0.366}&\textbf{0.378}&\textbf{0.353}&0.360&\textbf{0.353}&\textbf{0.353}
\end{tblr}\end{table}

\textbf{Comaprison with existing methods}
In evaluating the effectiveness of the Gaussian loss-weighted sampler, we compare it with existing methods such as InfoBatch (Table \ref{tab:multivariate-comparison-infobatch}). Using K-U-Net\_Linear\_1111, a multiplayer non-linear model, the Gaussian Loss Weighted sampler improves MSE for $4\%$ and Top5-MMSE for $1.7\%$ by dynamically calibrating the emphasis on learnable samples. This is contrasted against methods like InfoBatch which prunes easy samples and degrades MSE for $0.4\%-0.7\%$ and Top5-MMSE for $0.7\%-1.6\%$. Using K-U-Net\_Linear\_0000, a purely linear model, Gaussian methods degrade $0.1\%$ while Infobatch degrades  $0.3\%-0.5\%$. The results indicate that the Gaussian Loss Weighted sampler enhances model accuracy, making it a highly potential candidate for complex datasets where the other methods may be challenged.

\textbf{Comaprison across models}
Our study extends to comparing the impact of the Gaussian Loss Weighted sampler across various neural network architectures, including Channel Aligned Robust Blend Transformer (CARD)\citep{xue2024card} and PatchTST\citep{Nie-2023-PatchTST}. This cross-model analysis reveals how the sampler's performance adapts to different types of model (Table \ref{tab:multivariate-abalation_card_patchtst}). For CARD, the sampler effectively deceases prediction MSE error for $2.6\%$. In the case of PatchTST, it deceases the prediction error for $2.3\%$ . Meanwhile, the sampler improves the extreme performance by reducing the Top5-MMSE of CARD and PatchTST for $1.4\%$ and $1.8\%$ during the training. The experiments highlight the sampler’s adaptivity and its potential to bring consistent performance improvements across diverse machine learning models.

\begin{figure}[htb]
\centering

\begin{subfigure}[b]{.475\columnwidth}
\includegraphics[width=\columnwidth]{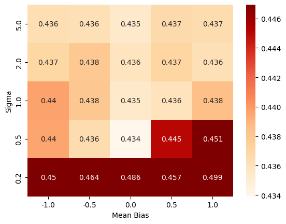} 
  \caption{ETTh1}
\end{subfigure}
 \hfill
\begin{subfigure}[b]{.475\columnwidth}
\includegraphics[width=\columnwidth]
{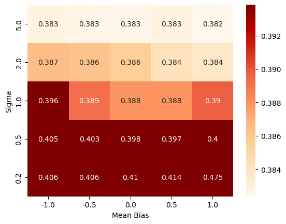}
  \caption{ETTh2}
\end{subfigure}

\vspace{10pt}

\begin{subfigure}[b]{.475\columnwidth}
\includegraphics[width=\columnwidth]
{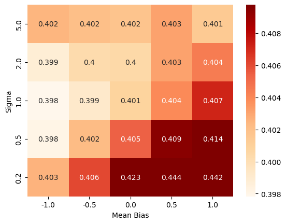}
  \caption{ETTm1}
\end{subfigure}
 \hfill
\begin{subfigure}[b]{.475\columnwidth}
\includegraphics[width=\columnwidth]
{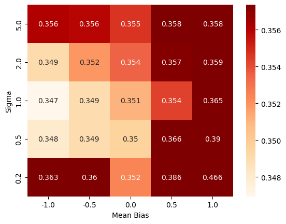}
  \caption{ETTm2}
\end{subfigure}

\vspace{10pt}
  \caption{Grid search of $\mu$ and $\sigma$ on ETT Dataset for forecasting task with lookback window $L=720$ and prediction horizon $T=720$}
  \label{fig:params_search}
\vspace{15pt}

\end{figure}

\textbf{Study of parameters $\mu$ and $\sigma$ }
In this study, we examine user-defined parameters mean bias (\(\mu\)) and standard deviation (\(\sigma\)) of the Gaussian distribution to gain a deeper understanding of their impact on the performance of learning algorithms. The standard deviation (\(\sigma\)), which controls the spread of the distribution, plays a critical role in determining how broadly the learning model considers the data around the mean (\(\mu\)), the central tendency. By manipulating these parameters, we can adjust the model's sensitivity to the importance of different samples during the training phase. This examination helps in fine-tuning the Gaussian Loss Weighted Sampler, optimizing it for specific datasets by balancing the focus between commonly occurring data points and outliers, thus potentially enhancing model robustness and accuracy. Empirically, setting mean bias $\mu$ to negative values(i.e., -0.5, -1) can help reduce the influence of outliers. In contrast, setting mean bias $\mu$ to positive values can make the model more proactive in learning rare outliers (Figure \ref{fig:params_search}).

\section{Conclusion}
We present the Gaussian loss-weighted resampling method, a novel method for training acceleration and robust regression. Gaussian Loss-weighted resampling shows its strong robustness on various datasets for time series forecasting tasks, achieving quality improvements and training acceleration. Gaussian Loss weighted resampling method improves $1\%-4\%$ MSE score compared to state-of-the-art methods, thus being practical for real-world applications. In the following work, we expect to test this method on other types of tasks such as classification and anomaly detection.


\bibliography{kun_gaussian}


\appendix
\section{Appendix}
\subsection{Detailed computation in Theorem 1}

 To solve for $ A $, $ B $, and $ C' $ in the given integral equations, let's begin by simplifying and comparing the integrands.

\textbf{Starting Equation}:
$$
C \int_0^{+\infty} \frac{1}{\sigma \sqrt{2\pi}} e^{-\frac{(x-\mu_x-\mu\sigma_x)^2}{2\sigma_x^2\sigma^2} + 2\lambda x} \, dx
$$

\textbf{Objective}:
We need to match this equation to a Gaussian integral in the form:
$$
C' \int_0^{+\infty} \frac{1}{B \sqrt{2\pi}} e^{-\frac{(x-A)^2}{2B^2}} \, dx
$$

\textbf{ Step 1:} 
\textit{Simplifying the exponent}

To match the given equations, first simplify the exponent in the original equation.

Given exponent:
$$
-\frac{(x-\mu_x-\mu\sigma_x)^2}{2\sigma_x^2\sigma^2} + 2\lambda x
$$

This can be rewritten in the form:
$$
-\frac{(x^2 - 2x(\mu_x + \mu\sigma_x) + (\mu_x + \mu\sigma_x)^2) - 4\sigma_x^2\sigma^2\lambda x}{2\sigma_x^2\sigma^2} 
$$

The nominator can be rewritten to:

$$
x^2 - 2x (\mu_x + \mu\sigma_x + 2\sigma_x^2\sigma^2\lambda  ) + (\mu_x + \mu\sigma_x + 2\sigma_x^2\sigma^2\lambda)^2+C''
$$

 where 
 $$
 C'' = -(\mu_x + \mu\sigma_x + 2\sigma_x^2\sigma^2\lambda)^2 + (\mu_x + \mu\sigma_x)^2
 $$
We can now rewrite the exponent term into : 

$$-\frac{1}{2\sigma_x^2\sigma^2}\left[(x-(\mu_x + \mu\sigma_x + 2\sigma_x^2\sigma^2\lambda ))^2 +  C'' \right]$$

\textbf{Step 2:} \textit{Factorize and Rearrange}

We rearrange the terms using constants $A$ and $B$ :

\begin{eqnarray}\label{eq:def_4}
A &=& (\mu_x + \mu\sigma_x + 2\sigma_x^2\sigma^2\lambda )\end{eqnarray}

 and 
 
\begin{eqnarray}\label{eq:def_5}
B&=&\sigma_x\sigma\end{eqnarray}

 Then the start equation becomes: 
$$
C \int_0^{+\infty} \frac{1}{\sigma \sqrt{2\pi}} e^{-\frac{(x-\mu_x-\mu\sigma_x)^2}{2\sigma_x^2\sigma^2} + 2\lambda x} dx
$$
$$=C \int_0^{+\infty}  \frac{\sigma_x}{B \sqrt{2\pi}} e^{-\frac{(x-A)^2}{2B^2} - \frac{C''}{2\sigma_x^2\sigma^2}} dx$$

\textbf{Step 3:} \textit{Calculate the Constant $C'$}

Now let us regroup the constant terms and $C$, we have :

$$C' \int_0^{+\infty}  \frac{1}{B \sqrt{2\pi}} e^{-\frac{(x-A)^2}{2B^2}} dx$$

where 

$$C' = C\sigma_x e^{- \frac{C''}{2\sigma_x^2\sigma^2}}$$

$$ = C\sigma_x e^{- \frac{-(\mu_x + \mu\sigma_x + 2\sigma_x^2\sigma^2\lambda)^2 + (\mu_x + \mu\sigma_x)^2}{2\sigma_x^2\sigma^2}} $$

\begin{eqnarray}\label{eq:def_6} &=&C\sigma_x e^{\frac{(\mu_x + \mu\sigma_x + 2\sigma_x^2\sigma^2\lambda)^2 - (\mu_x + \mu\sigma_x)^2}{2\sigma_x^2\sigma^2}} 
\end{eqnarray}


\end{document}